\documentclass[conference]{IEEEtran}
\IEEEoverridecommandlockouts

\usepackage{cite}
\usepackage{amsmath,amssymb,amsfonts}
\usepackage{algorithmic}
\usepackage{graphicx}
\usepackage{textcomp}
\usepackage{xcolor}
\def\BibTeX{{\rm B\kern-.05em{\sc i\kern-.025em b}\kern-.08em
    T\kern-.1667em\lower.7ex\hbox{E}\kern-.125emX}}

\usepackage{booktabs}
\usepackage{tikz}
\usetikzlibrary{calc}
\usepackage{xcolor}

\usetikzlibrary{math}

\usepackage[hyphens,spaces,obeyspaces]{url}
\usepackage{breakurl}
\usepackage[breaklinks,colorlinks,allcolors=black]{hyperref}

\tikzset{
  pics/targetmarker/.code={
    \def\a{0.5cm} \def\b{0.2cm} \def\L{1mm}
      
    \draw (0,0) ++({\a*cos(-210)},{\b*sin(-210)})
      arc[start angle=-210, end angle=30, x radius=5mm, y radius=2mm];

    \foreach \ang in {-160,-120,-80,-40,0} {%
      \pgfmathsetmacro{\cx}{\a*cos(\ang)}
      \pgfmathsetmacro{\cy}{\b*sin(\ang)}
      \pgfmathsetmacro{\nx}{\b*cos(\ang)}
      \pgfmathsetmacro{\ny}{\a*sin(\ang)}
      \pgfmathsetmacro{\s}{veclen(\nx,\ny)}
      \pgfmathsetmacro{\ux}{\nx/\s}
      \pgfmathsetmacro{\uy}{\ny/\s}
      \draw ({\cx - 0.5*\L*\ux},{\cy - 0.5*\L*\uy})
        -- ({\cx + 0.5*\L*\ux},{\cy + 0.5*\L*\uy});
    }
  }
}

\begin{document}

\title{Distant Object Localisation from Noisy Image Segmentation Sequences
\thanks{
This research was funded by the Research Council of Finland within projects DRONE4TREE (decision no. 359404), ML4DRONE (decision no. 357380), and Fireman (decision no. 346710) and with grants (no. 339730 and 362408). The FireMan project is funded under the EU’s Recovery and Resilience Facility, that promotes the green and digital transitions through research. This study has been performed with affiliation to the RCF Flagship Forest–Human–Machine Interplay—Building Resilience, Redefining Value Networks and Enabling Meaningful Experiences (UNITE) (decision no. 357908). 
}}
\author{\IEEEauthorblockN{Julius Pesonen}
\IEEEauthorblockA{Dept. of Remote Sensing and Photogrammetry \\
{Finnish Geospatial Research Institute}\\
{and Aalto University}\\
Espoo, Finland \\
julius.pesonen@nls.fi}
\and
\IEEEauthorblockN{Arno Solin}
\IEEEauthorblockA{ELLIS Institute Finland \\
{and Aalto University}\\
Espoo, Finland}
\and
\IEEEauthorblockN{Eija Honkavaara}
\IEEEauthorblockA{Dept.\ of Remote Sensing and Photogrammetry \\
{Finnish Geospatial Research Institute}\\
Espoo, Finland}

}

\maketitle

\begin{abstract}

3D object localisation based on a sequence of camera measurements is essential for safety-critical surveillance tasks, such as drone-based wildfire monitoring. Localisation of objects detected with a camera can typically be solved with specialised sensor configurations or 3D scene reconstruction. However, in the context of distant objects or tasks limited by the amount of available computational resources, neither solution is feasible. In this paper, we show that the task can be solved with either multi-view triangulation or particle filters, with the latter also providing shape and uncertainty estimates. We studied the solutions using 3D simulation and drone-based image segmentation sequences with global navigation satellite system (GNSS) based camera pose estimates.  The results suggest that combining the proposed methods with pre-existing image segmentation models and drone-carried computational resources yields a reliable system for drone-based wildfire monitoring. The proposed solutions are independent of the detection method, also enabling quick adaptation to similar tasks. Code is available at: \url{https://fgi\_nls.gitlab.io/public/distant-localisation}

\end{abstract}

\section{Introduction}
\label{sec:intro}

This work explores solutions for locating very distant objects from a series of camera-based detections from known locations and orientations. At a glance, the problem of locating target objects based on multiperspective imagery seems well-addressed. 
However, the task poses unique problems when applied to objects that might be located multiple kilometres away. 
Typical solutions to such tasks include specific sensor configurations relying on stereo cameras or time-of-flight sensors, such as lidar.
Unfortunately, at such scales, stereo cameras would require huge baselines, and time-of-flight sensors would become unreliable due to their 3D spatial resolution worsening cubically by distance. 
Alternatively, full camera-based 3D reconstruction methods have been applied to such problems, but creating a 3D model of such a vast region becomes computationally inefficient when the goal is to determine the position of a single target or those of only a few target objects.

\begin{figure}[t]
  \centering
  \begin{tikzpicture}[inner sep=0]

    \node[minimum width=.33\columnwidth,minimum height=.2\columnwidth] at (4,4.45) {\includegraphics[width=.35\columnwidth]{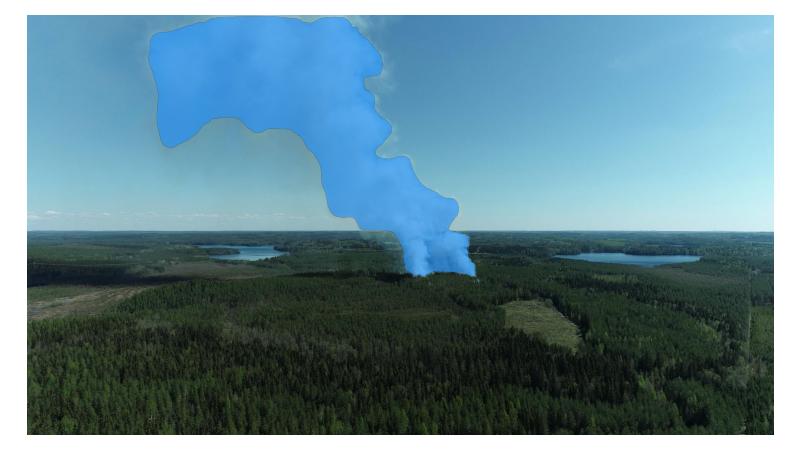}};    
    \node[minimum width=.33\columnwidth,minimum height=.2\columnwidth] at (7,4.45) {\includegraphics[width=.35\columnwidth]{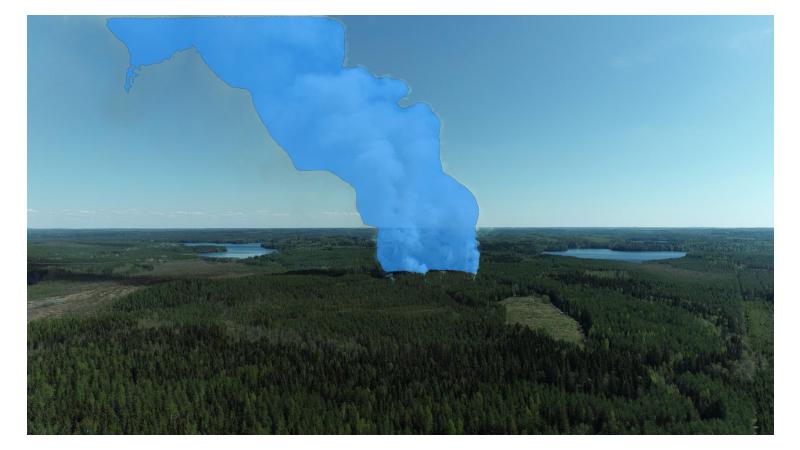}};
  
    \node[anchor=south west] (forest) at (0,0) {\includegraphics[width=\columnwidth]{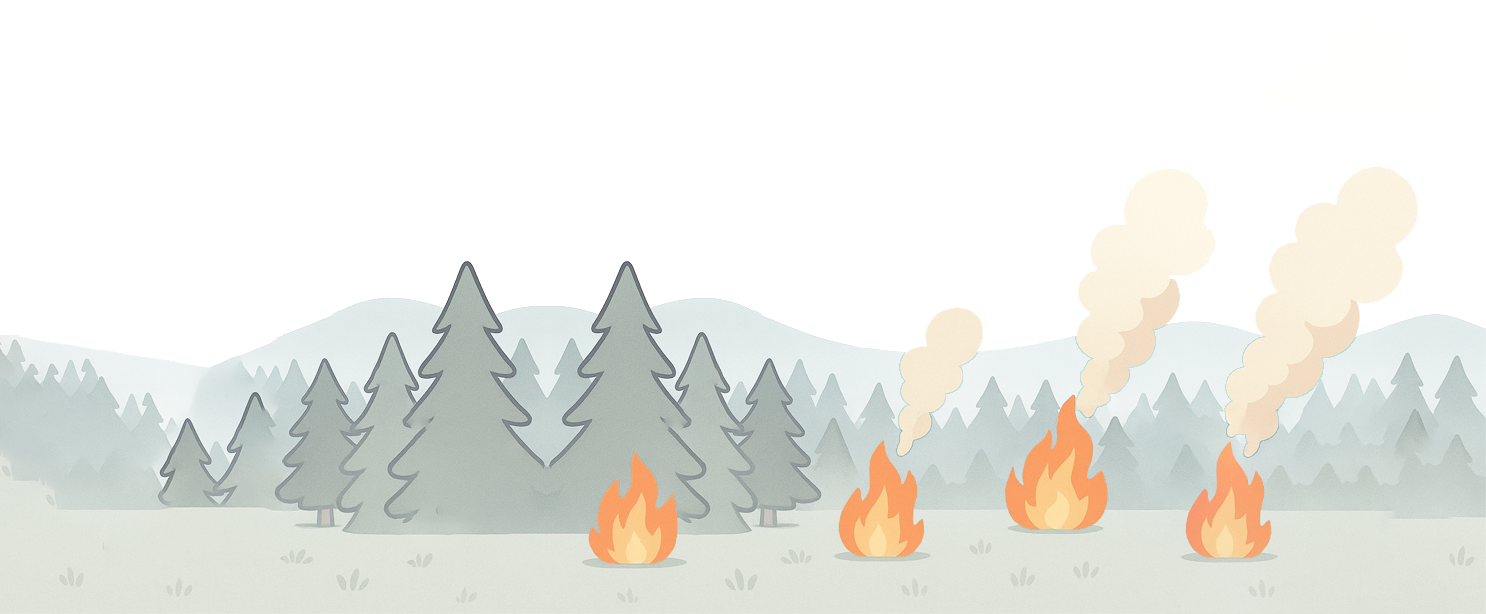}};
    \node (drone) at (2,4) {\includegraphics[width=.4\columnwidth]{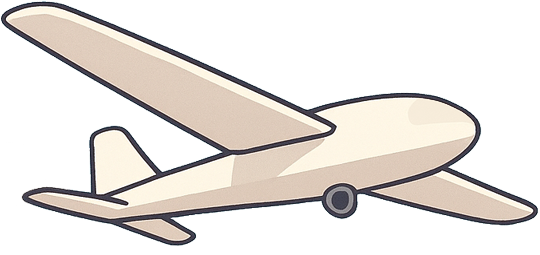}};

    \fill[black!30,opacity=.10] ($(drone.center) + (5mm,-5mm)$) -- ++(-40:4) -- ($(drone.center) + (5mm,-5mm) + (-10:5)$);
    \draw[black!80,densely dashed] ($(drone.center) + (5mm,-5mm)$) -- ++(-40:4);
    \draw[black!80,densely dashed] ($(drone.center) + (5mm,-5mm)$) -- ++(-10:5);

    \begin{scope}[shift={(6.3,.5)}]
       \pic {targetmarker};
    \end{scope}
    
    \begin{scope}[shift={(7.35,.35)}]
       \pic {targetmarker};
    \end{scope}
    
    \node[font=\scriptsize] at (2,2.75) {Moving camera};
    \node[font=\scriptsize,align=right,anchor=east,inner sep=1pt] at (\columnwidth,.75) {Estimated\\positions};
    
  \end{tikzpicture}
  \caption{We propose a hybrid approach for localising distant objects/events (such as wildfire smoke) from sequences of frames and GNSS-estimated poses from a moving RGB camera (example frames with masked smoke shown on top-right).}
  \label{fig:teaser}
\end{figure}

The motivation for this work originates from drone-based wildfire detection, in which the position of the drone-carried camera is estimated using GNSS measurements and known dynamics of the drone camera setup. Our earlier work~\cite{Pesonen_2025_WACV} showed that wildfire smoke can be detected from almost ten kilometres away using only drone-carried resources. Pairing the segmentation model with a lightweight target localisation method would enable fully on-board wildfire detection and localisation. 
This means that the wildfire detection system could be deployed in areas of poor telecommunication where cloud-based computing can not be relied on. 
The sketch in Figure~\ref{fig:teaser} illustrates a use case of a UAV scanning for wildfires with masked RGB images.

Besides the perception distance, smoke clouds as the perceived targets pose unique challenges due to their practically unlimited variety of shapes. 
To get a reliable estimate of the possible wildfire positions in ground coordinates, it would be beneficial to get a prediction of the shape of the smoke cloud and any possible uncertainties presented by the perception framework. 
A variety of sources for uncertainties are presented in the task, as even small errors in the camera pose estimation cause major differences in perceived positions as the sensing distances grow larger. 
With far enough objects, it also becomes practically impossible to see a large distant object from such a variety of angles that it could be perfectly enclosed in a single position. 
Thus, it's beneficial to use methods which could present the full region of possible locations for the target object.

To study the possible solutions, we created a simple simulation where target objects are simplified as cubes. 
We also evaluated the presented methods using two sequences of drone footage where the target is first presented by a telecommunication mast and then by a smoke cloud originating from an industrial chimney. 
Using these experiments, the study shows that the 3D centre point of the target can be determined both using robust multi-view triangulation or particle filters. The latter of which is also capable of providing a rough estimate of the target object's shape and the resulting localisation uncertainty.

\section{Background}
\label{sec:rw}

The task at hand can be thought of as a form of camera-based multi-view 3D reconstruction. The earlier developments of 3D object reconstruction from multi-view imagery have been well documented by Hartley and Zisserman~\cite{Hartley2004}. These methods relied on point correspondence, well-estimated camera parameters, and bundle adjustments. Later developments, such as COLMAP~\cite{schoenberger2016sfm} and 3D Gaussian splatting~\cite{kerbl3Dgaussians}, have greatly improved the structure from motion map generation and 3D modelling, respectively. Unfortunately, the methods still require finding a large number of point correspondences between frames, making them computationally heavy for iterative real-time target localisation using edge devices, which is desirable in, for example, aerial robotic applications. 

The 3D reconstruction methods have been poorly studied in the context of smaller and more distant objects. In such situations, any noise in the camera pose estimation causes larger errors in the final reconstruction. This suggests that robust techniques such as Bayesian filters could offer feasible solutions. Besides the camera pose estimation errors, typical modern camera-based sensing solutions leverage neural network models for detecting key features, which complicates keypoint matching between frames in scenarios where we do not want the target object to be reduced to a single point in 3D space. 

Even though the observed object, in this study, is assumed to be static, the problem relates to object tracking, as we're interested in obtaining a position for an object in 3D space from a set of camera observations. Like 3D reconstruction, camera-based object tracking has been studied since the dawn of time, and a survey covering the earlier developments has been written by Yilmaz et al.~\cite{yilmaz2006object}. More recent surveys also consider neural network-based approaches and list a large number of methods, evaluation metrics, and datasets for the study of camera-based object tracking~\cite{CIAPARRONE202061,CHEN2022103508,awal2023particle} even for real-time scenarios~\cite{Kalake2021AnalysisBO}. However, even these more recent surveys only consider metrics based on image-based labels, failing to consider the 3D localisation errors. This limitation also applies to a survey on video object segmentation and tracking~\cite{yao2020video}. Multitarget detection and tracking from a monocular camera has also been studied specifically in the context of drones, but again, the tracking accuracy has only been evaluated in the camera plane~\cite{7759733}.

Still, the 3D object tracking problem is not entirely devoid of resources. One example where the problem has been studied is in autonomous driving, in which benchmarks and datasets such as the KITTI dataset~\cite{Geiger2012CVPR} and NuScenes~\cite{caesar2020nuscenes} have enabled major progress. The autonomous driving datasets, however, typically include other sensor modalities such as lidar or stereo cameras, which are not feasible to be used in more distant localisation scenarios. Some studies consider only the monocular camera view~\cite{wang2023exploring}, but the focus on nearby objects still makes the problem statement very different. Other examples of works in the 3D space include tracking of people in small-scale indoor scenes with static camera views, also using particle filters~\cite{lopez2007multi,salih20113d}, and small object tracking with a single static $360^\circ$ field of view (FOV) camera~\cite{taiana20073d}.

In the multisensor context, the use of cameras has been slightly more common as well. 
However, in the multicamera scenario, the problem is inherently different from that of a single camera, due to the possibility of using the discrepancies between the multiple sensor locations with simultaneous observations. Besides, the metrics in the literature have been focused on the 2D labels as with individual cameras~\cite{AMOSA2023126558}.


Filter-based methods have also been applied to drone-based tasks, and benchmarks have been presented for localising human or vehicle targets based on separate views from multiple drones~\cite{zhu2020multi, Liu2023RobustMM}. 
The work by Liu and Zhang~\cite{liu2021vision} presents a very similar task to the one at hand, where objects such as cars were tracked and localised from drone-captured imagery with a combination of a neural network and a particle filter. However, the localisation was simplified by assuming a flat ground and a triangular relationship between the ground plane, the target, and the drone-based camera. In addition, the scale of the task is still very different from what is of interest in this study, meaning the range of multiple kilometres.

\section{Materials and Methods}

We propose estimating the distant target position using a particle filter. 
We used simulation and real drone captured footage with GNSS-measured camera positions to evaluate and compare the filter's performance to that of multi-view triangulation. 

\subsection{Problem Setup}

To study the problem in detail, we defined a simulation in which a target object is projected onto a camera plane using a pinhole camera model. The simulation was flexible and allowed quick testing of different targets, distances, and noise variations. 

The simulated target was defined as a three-dimensional cube for simplicity. The cube was defined by its eight corners, which were used to project the cube into the simulated two-dimensional camera image, where a single point projection was computed as
\begin{equation}
    \mathbf{y} = \mathbf{K M x},
\end{equation}
where $\mathbf{y}$ is the projected point in homogeneous coordinates, $\mathbf{K}$ is the intrinsic camera matrix, $\mathbf{M}$ is the extrinsic camera matrix, and $\mathbf{x}$ is the original 3D point in homogeneous coordinates. The homogeneous coordinates use an extra dimension to simplify the matrix operations, such that $\mathbf{y} = (y_1, y_2, y_3)^\top$ and $\mathbf{x} = (x_1, x_2, x_3, x_4)^\top$, where the pixel coordinates of the projection are: 
\begin{equation}
    \mathbf{\hat{y}} = (y_1/y_3, y_2/y_3)^\top,
\end{equation}
and the corresponding 3D world coordinates are:
\begin{equation}
\mathbf{\hat{x}} = (x_1/x_4, x_2/x_4, x_3/x_4)^\top.
\end{equation}

The intrinsic and extrinsic camera matrices, $\mathbf{K}$ and $\mathbf{M}$, describe the physical parameters of the camera. $\mathbf{K}$ is a $3{\times}3$ matrix:
\begin{equation}
    \mathbf{K} = \begin{bmatrix}
        f_x & 0 & c_x \\
        0 & f_y & c_y \\
        0 & 0 & 1 \\
    \end{bmatrix},
\end{equation}
where $f_x$ and $f_y$ describe the focal length of the camera in terms of pixels, and $c_x$ and $c_y$ describe the position of the principal point of the camera. 
The extrinsic matrix is a $3{\times}4$ matrix consisting of a $3{\times}3$ rotation matrix, $\mathbf{R}$, and a $3{\times}1$ translation component, $\mathbf{t}$, corresponding to each 3D translation axis:
\begin{equation}
    \mathbf{M} = 
    \begin{bmatrix}
        \mathbf{R} & \mathbf{t} \\ 
    \end{bmatrix}
    = 
    \begin{bmatrix}
        r_{11} & r_{12} & r_{13} & t_{x} \\
        r_{21} & r_{22} & r_{23} & t_{y} \\
        r_{31} & r_{32} & r_{33} & t_{z} \\
    \end{bmatrix}.
\end{equation}

After the projection, the image coordinates were discretised to integer values corresponding to camera pixel coordinates. This was essential to simulate the loss of information that results from using both cameras and segmentation models with a limited number of pixels, in a distant observation scenario. To produce the actual simulated segment from the pixel projections, a convex hull was used to obtain the area that covers the whole view of the cube in the camera frame. This convex hull then represented the \emph{perfect} segmentation result. 

To simulate the noise caused by non-perfect camera pose estimation, we injected the camera extrinsics corresponding to each frame with a random amount of translation noise, denoted $\mathbf{\nu_t}$, in each coordinate axis and rotated the matrix over each axis separately, again with a random amount, denoted $\mathbf{N_r}$. This resulted in a noisy extrinsic matrix $\mathbf{M_\nu}$, defined as:
\begin{multline}
    \mathbf{M_\nu} = 
    \begin{bmatrix}
        \mathbf{N_{rx} N_{ry} N_{rz} R} & \mathbf{t + \nu_t} \\ 
    \end{bmatrix} \\
    = 
    \begin{bmatrix}
        r_{\nu11} & r_{\nu12} & r_{\nu13} & t_{x} + \nu_{tx}\\
        r_{\nu21} & r_{\nu22} & r_{\nu23} & t_{y} + \nu_{ty}\\
        r_{\nu31} & r_{\nu32} & r_{\nu33} & t_{z} + \nu_{tz}\\
    \end{bmatrix},
\end{multline}
where $\mathbf{N_{rx}}$, $\mathbf{N_{ry}}$, and $\mathbf{N_{rz}}$ correspond to the separately drawn rotation noise matrices, $r_{\nu n}$ to each resulting noisy rotation element, and $\nu_{tx}$, $\nu_{ty}$, and $\nu_{tz}$ to each separately drawn translation noise element.

The false positives in the simulated segments were generated by defining random rectangular sections of the image and setting these pixels equal to the real positive segments. These false-positive rectangles were defined by their height and width in pixel dimensions. They were generated randomly for each image, based on a false positive rate $\rho_{FP}$ and the maximum number of false positives $\mathbf{Max_{FP}}$. Correspondingly, the false positives were removed from the following frames based on a false positive dismissal rate $\delta\rho_{FP}$. Unless the false positive was removed, it was kept for the subsequent images. 

The false negatives were simulated in two ways. First, by simply setting all the target segment pixels to zero, equalling the value of the background or by setting only some randomly selected section of the target pixels to zero. While the false positives and partial false negatives were kept in consequent frames until dismissal defined by another random draw, the false negatives were generated independently for each frame. Thus, the false negative appearance and disappearance were defined by the false negative rate $\rho_{FN}$, by the partial false negative rate $\rho_{PFN}$, and the partial false negative dismissal rate $\delta\rho_{PFN}$.

\subsection{Multi-view Triangulation}

The simplest solution to the problem is presented by multi-view triangulation. 
The method allows determining the centre point of the target object by solving a least squares estimate of a point between the camera rays obtained from different viewing points. 
In practice, this means generating a 3D ray based on each segmentation frame and camera matrix, solving a direct linear transform (DLT), and finding the least squares position. 

In practice, we implemented multi-view triangulation for the task by reducing the segments into individual pixel coordinates by taking their centre positions and solving the DLT. 
The DLT was solved by first generating, for each camera pose and the corresponding segment centre, two linear equations that were used to ensure that the 3D point lies in the correct horizontal and vertical planes, respectively:
\begin{equation}
    y_3 (\mathbf{M_1 \hat{x}}) - y_1 (\mathbf{M_3 \hat{x}}) = 0,
\end{equation}
\begin{equation}
    y_2 (\mathbf{M_3 \hat{x}}) - y_3 (\mathbf{M_2 \hat{x}}) = 0,
\end{equation}
where $\mathbf{M_1}$, $\mathbf{M_2}$, and $\mathbf{M_3}$ denote the rows of the camera matrix $\mathbf{M}$.
Stacking all the equations provided a system of linear equations, $\mathbf{A\hat{x}}=0$, which was minimised using the singular value decomposition of $\mathbf{A}$. 
The minimised solution for $\mathbf{\hat{x}}$ was the 3D target position estimate in homogeneous coordinates.

To make the triangulation more robust to outliers, which are abundant in the studied scenario, we used the random sample consensus (RANSAC) algorithm. 
The RANSAC discards the outlier camera positions and segments by evaluating a reprojection error 
\begin{equation}
    E_{reproj} =  ||\mathbf{y} - \mathbf{\bar{y}}||,
\end{equation}
where $\mathbf{\bar{y}}$ is the reprojection of a candidate position $\mathbf{\hat{x}}$ computed by the standard DLT of two randomly sampled camera positions and segments. 
$E_{reproj}$ was minimised until a sufficient number ($\ge80\%$) of observations were considered inliers ($E_{reproj}<2$). If less than 80\% were inliers, the configuration with the highest number of inliers was used.
Finally, all the inlier positions were used to compute a final estimate using the standard DLT. 
The algorithm thus discarded outlier segments, finding a more reliable solution. 

\subsection{Particle Filter}

The multi-view triangulation was, in this scenario, only able to determine the centre of position of the target object, which is unreliable for a practical task such as wildfire localisation. As such, we also show that the same estimates can be produced using a particle filter, which simultaneously provides rough object shape and uncertainty estimates.

\paragraph{\textbf{Initialisation}}
We defined the filter such that the particles were initialised uniformly along a line in 3D space corresponding to the camera ray of the first observation. The limits of the distribution were set to 50 and 30~000 metres from the camera. 

\paragraph{\textbf{Prediction step}} The step was first taken immediately after the initialisation. At each prediction step, the particles were injected with independent Gaussian noise defined by the individual particle variation $\mathbf{\sigma_p}$, based on a domain-specific constant. This meant that after each prediction step, the full distribution could be expressed as a new combination of $N$ three-dimensional Gaussians, where $N$ is the number of particles. 
This distribution serves as the target position prediction of the model. 

\paragraph{\textbf{Update step}} The update was only done given that positive pixels were observed. In the update step, the pixel distances between the projected particles and the positive observations were compared. Each particle, projected inside the camera frame, was assigned a weight, $\mathbf{\omega}_{p}$, relative to the distance from the positive pixels such that:
\begin{equation}
    \mathbf{\omega}_{p} = e^{-\min((\mathbf{obs}-\mathbf{p_{proj}})^2)},
\end{equation}
where $\mathbf{obs}$ is an array of the positive observations and $\mathbf{p_{proj}}$ is an array where each element presents the same individual projected particle in the pixel coordinates, with the number of elements matching those of the observations. 

\paragraph{\textbf{Resampling}} We used the bootstrap implementation of the particle filter, meaning that the resampling was performed after each update step, by using the weights as the probabilities of drawing the corresponding points from the set of particles. The same number of particles as the previous set was drawn in this manner, resulting in a new set where the highest weighted particles were repeated multiple times. This resulting distribution was then used in the next prediction step, where particles were first updated randomly, effectively doing Gaussian kernel smoothing for the distribution, before projecting them again to the camera frame. We chose the bootstrap version of the particle filter due to better initial results.

\subsection{Metrics}

The performance of the proposed method was quantified, most importantly, by the root mean square error (RMSE), which was computed between the predicted particles, $\mathbf{p}_i$, and the mean of the target location, $\mathbf{m}_t$:
\begin{equation}
    \mathrm{RMSE} = \sqrt{\dfrac{\sum_{i=1}^{n_p}(\mathbf{p}_{i}-\mathbf{m}_{t})^2}{n_p}},
\end{equation}
where $n_p$ is the number of particles. This is equal to the Euclidean distance between the predicted and ground truth means.

Another measure used for evaluating the particle filter, specifically in the simulation, was the ratio of particles which fell in the target object region in the 3D space. The ratio was computed as $ratio = n_{in} / n_p$, where $n_{in}$ and $n_p$ were the number of particles inside the target object and the total number of particles, respectively. 
The ratio enabled quantitatively observing how well the particle filter distribution converges to that of the actual target object. 

For both metrics, from the simulations, we report the mean between 200 and 1000 metres of camera translation, and for the RMSE, we also report the minimum obtained value during the test. The first 200 metres were dismissed due to being mostly dependent on the initialisation, and 1000 metres was the maximum translation in the experiments. For all results, the metrics were computed over an average of ten simulations to reduce noise caused by the randomness of the experiments. For the particle filter, the multiseed configuration was also used when evaluating with empirical data due to the method's dependency on the pseudo-randomly varied particles.

\subsection{Empirical Data}

To validate that the performance of the proposed method holds for real-world applications, we tested the system on two drone-captured video sequences. 
The position of the drone in each was recorded using GNSS, and the approximate geolocations of the targets were known. 

To capture the first sequence with a telecom mast target, we used a DJI Matrice 350, equipped with an AR0234 camera and an Applanix APX-15 UAV GNSS. An NVIDIA Jetson Orin NX was used to record the data. 
We performed no calibration for the camera or the GNSS. 
This means that for the camera, only the parameters provided by the camera manufacturer were used~\cite{arducamAR0234Arducam}. The camera was used to record at full HD (1080 by 1920 pixels) resolution with a horizontal FOV of 90$^\circ$ and vertical FOV of 50.625$^\circ$. This means that the intrinsic parameters were set as $f_x = 1200$, $f_y = 1200$, $c_x = 960$, and $c_y = 540$. The lens was assumed to cause no distortions. 
The GNSS antenna was mounted approximately 30 centimetres away from the camera with an external IMU attached to the camera. Neither the boresight nor the lever arm of the mounting was taken into account for the experiments. This means that the camera pose estimation had at least a systematic error in the range of tens of centimetres in addition to any sensor-caused errors. The manufacturer reported RMSE for the position is 0.02 to 0.05 metres, for the roll and pitch 0.025 degrees, and for the heading 0.080 degrees when using differential GNSS~\cite{APX15UAV}.
The maximum camera translation in the sequence was approximately 250 metres, with the mean distance to the target at approximately 700 metres. 

The second sequence, with an industrial smoke cloud as the target, was captured using a DJI Mini 3. 
Again, the video was captured in full HD resolution but with a slightly different FOV.
The DJI Mini 3 horizontal FOV of 82.1° corresponds to a focal length $f_x = f_y = 1102.41$.
The reported hovering accuracy of the GNSS was 0.5 metres vertically and 1.5 metres horizontally~\cite{DJIMini3}. 
The angular error ranges of the camera pose estimate were not reported. 
As for the first sequence, no additional calibration was performed for the system before recording the data.
The maximum camera translation in the sequence was approximately 230 metres, with the mean distance to the target at 1770 metres. 

\begin{table*}[t]
    \centering
    \caption{Simulated filter position estimation results with different noise and target configurations.}
    \begin{tabular}{p{1.cm}p{1.cm}p{1.cm}p{1.cm}p{1.cm}p{1.cm}p{1.cm}p{1.cm}p{1.cm}p{1.cm}p{1.4cm}p{1.cm}}
       \toprule
       Method & Max $\nu_{rot}$ ($\circ$) & Max $\nu_{t}$ (m) & $\rho_{FP}$ & $\delta\rho_{FP}$ & Max $FP$ & $\rho_{FN}$ & $\rho_{PFN}$ & $\delta\rho_{PFN}$ & RMSE min (m) & RMSE 200-1k (m) & Ratio 200-1k \\ 
       \midrule
       MVT & 0 & 0 & 0 & - & 0 & 0 & 0 & - & 1.06 & 3.95 & - \\  
       MVT & 0.5 & 0.1 & 0 & - & 0 & 0 & 0 & - & 1.08 & 3.79 & - \\  
       MVT & 0.5 & 0.1 & 0.1 & 0.2 & 3 & 0 & 0 & - & 1.08 & 147.84 & - \\
       MVT & 0.5 & 0.1 & 0.1 & 0.2 & 3 & 0.1 & 0 & - & 1.32 & 277.88 & - \\
       MVT & 0.5 & 0.1 & 0.1 & 0.2 & 3 & 0.1 & 0.1 & 0.2 & 1.54 & 257.67 & - \\

       RMVT & 0 & 0 & 0 & - & 0 & 0 & 0 & - & 1.06 & 3.92 & - \\  
       RMVT & 0.5 & 0.1 & 0 & - & 0 & 0 & 0 & - & 1.25 & 3.88 & - \\  
       RMVT & 0.5 & 0.1 & 0.1 & 0.2 & 3 & 0 & 0 & - & 1.21 & 3.66 & - \\
       RMVT & 0.5 & 0.1 & 0.1 & 0.2 & 3 & 0.1 & 0 & - & 1.26 & 3.21 & - \\
       RMVT & 0.5 & 0.1 & 0.1 & 0.2 & 3 & 0.1 & 0.1 & 0.2 & 1.20 & 3.94 & - \\

       PF & 0 & 0 & 0 & - & 0 & 0 & 0 & - & 9.44 & 44.83 & 0.15 \\  
       PF & 0.5 & 0.1 & 0 & - & 0 & 0 & 0 & - & 9.57 & 45.67 & 0.15 \\
       PF & 0.5 & 0.1 & 0.1 & 0.2 & 3 & 0 & 0 & - & 10.99 & 46.97  & 0.15 \\
       PF & 0.5 & 0.1 & 0.1 & 0.2 & 3 & 0.1 & 0 & - & 13.04 & 46.86 & 0.15 \\
       PF & 0.5 & 0.1 & 0.1 & 0.2 & 3 & 0.1 & 0.1 & 0.2 & 17.87 & 63.97 & 0.15 \\
       \bottomrule
    \end{tabular}
    \label{tab:simple_results}
\end{table*}

The images of the first sequence were post-processed using a horizontal Sobel operation with image erosion and dilation.
This way, we obtained a sequence of binary segments with a known target location and estimated camera positions, corresponding to a drone-based target localisation task. The known target geolocation also enabled measuring the quality of the localisation in ground coordinates.

From the second sequence, the smoke cloud was segmented using SAM 3~\cite{carion2026sam} with the text prompt 'smoke'. The model recognised separate instances of smoke clouds, but only the nearest visible smoke cloud was used in the evaluation. The other segments were discarded. 
The errors for the test were measured from the top of the industrial chimney. As the smoke cloud was moving away from the chimney, the expected error was non-zero. Based on visual estimation, errors from 50 to 200 metres were expected.

\section{Experiments and Results}

\begin{figure*}[t]
\newlength{\figheighta}
\newlength{\figwidtha}
\setlength{\figwidtha}{0.125\textwidth}
\setlength{\figheighta}{.5625\figwidtha}
\begin{tikzpicture}
  \foreach \i in {1,2,3,4,5,6,7,8} {
    \node[minimum width=\figwidtha,minimum height=\figheighta] at (\i*\figwidtha,0) {\frame{\includegraphics[width=\figwidtha]{figs/sim_frames/noiseless_single_inverse/frame_\i.jpg}}};
    \node[minimum width=\figwidtha,minimum height=\figheighta] at (\i*\figwidtha,-\figheighta) {\frame{\includegraphics[width=\figwidtha]{figs/sim_frames/noisy_single_inverse/frame_\i.jpg}}};
    \tikzmath{\k = (\i-1)*125;}
    \node[font=\footnotesize] at (\i*\figwidtha, 0.65\figheighta) {$\k$~m};
    }
\end{tikzpicture}
    \includegraphics[width=\textwidth]{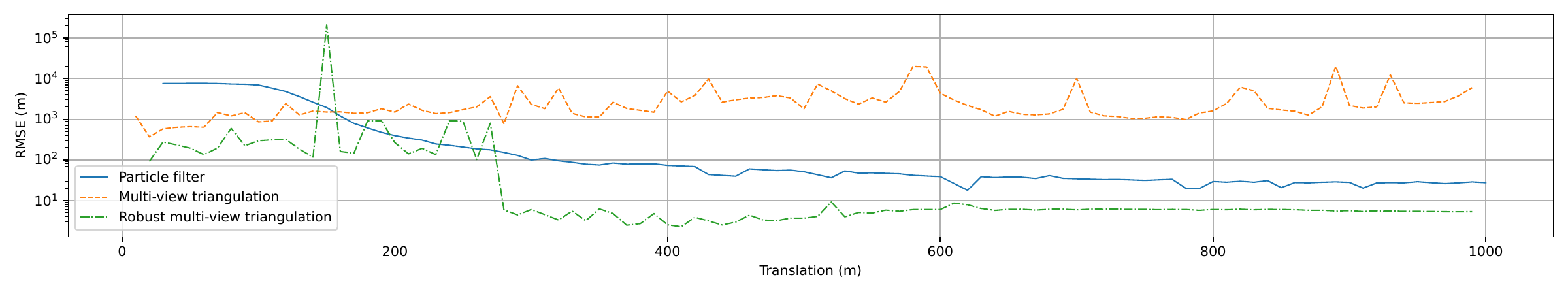}
    \caption{Noisy single target simulation results. From top to bottom: Simulated camera translation from the start of the sequence, noiseless single target simulation sample frames, fully noisy simulation samples, RMSEs of the single target simulation experiments over the camera translation with a logarithmic RMSE axis. The plot highlights the smooth convergence of the particle filter predictions compared to those of the multi-view triangulation.}
    \label{fig:singletarget}
\end{figure*}

Our experiments demonstrate that the proposed particle filter can estimate distant target positions as effectively as multi-view triangulation. 
The evaluation was done using a simulation, visualised in Figure~\ref{fig:singletarget}, and two drone-captured image segmentation sequences of a telecommunication mast shown in Figure~\ref{fig:rw_rmse} and that of an industrial chimney smoke cloud shown in Figure~\ref{fig:smoke_sequence}.
With our results, we also demonstrate how the particle filter is able to distinguish the localisation uncertainty and provide a coarse estimate of the target object's shape and size.

\subsection{Simulation Study}

The simulation offered a possibility to study the method extensively by enabling testing in an unlimited number of different scenarios varying by, for example, the camera trajectories, noise levels, and observation distances. Here, we present the simulation results in an order of increasing complexity. Since the number of possible variations is infinite, we tried to limit the results to scenarios which could offer the most insight into the method's performance in the expected real-world tasks as well as its robustness to the different noise sources. The quantitative simulation results are presented in Table~\ref{tab:simple_results} with the corresponding simulation experiments explained in the following paragraphs. The convergence of the filter in the corresponding experiments is visualised in Figure~\ref{fig:singletarget}.

The camera intrinsic parameters were set as $f_x = 1200$, $f_y = 1200$, $c_x = 960$, and $c_y = 540$, to imitate the camera used in the first empirical test. The discretisation caused by the camera pixels was taken into account in all simulations. 

The simulated perpendicular observation trajectory presents the optimal and simplest target observation scenario. Without noise, this presents an exceedingly optimal situation, and we mainly used it to confirm that the models performed as expected and to analyse the minimum uncertainty and errors which could theoretically be obtained when locating distant visual targets. We performed these tests with a single 100 by 100 by 100 metre target, with the target position, $\mathbf{tp} = (x, y, z)^\top = (500, -200, 2000)^\top$, in metres. The trajectory of the camera positions, $\mathbf{cp}$, started from $\mathbf{cp} = (0, 0, 0)^\top$ and continued linearly to $\mathbf{cp} = (1000, 0, 0)^\top$, in metres. The camera angle remained stationary in the simulation, and it was defined such that when the target and camera positions' x-coordinates were equal, the target was projected in the horizontal middle axis of the camera frame. The pitch angle of the camera was set such that the horizon was in the middle of the camera frame's vertical axis. 

Even with perfectly noiseless segmentation and an optimal trajectory, the camera requires some translation for the particle filter position estimates to converge to the correct position.
However, the estimates quickly converge to a position with only approximately $<$5\% relative RMSE to the target distance of two kilometres. 
As expected, the multiview triangulation already finds the correct position with only a few frames when no noise is present, and the only errors are caused by the pixel discretisation.

\begin{figure}[!ht]
    \centering
    \includegraphics[width=0.98\linewidth]{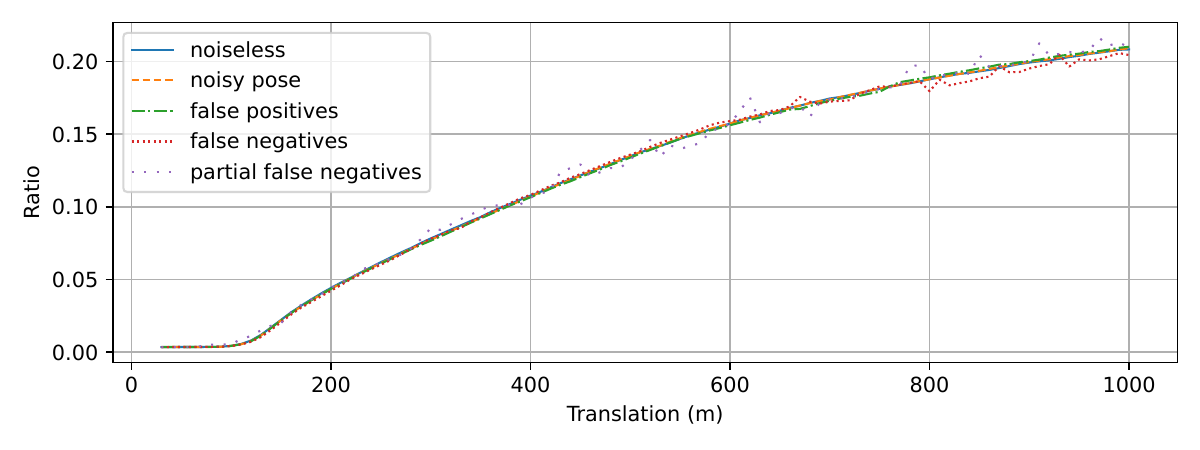}
    \caption{Convergence of the particle ratio over the camera translation in simulation in different noise scenarios. The types of noise are additive as in Table~\ref{tab:simple_results}. The almost equal ratio curves show that the particle filter converges towards the right region despite the various noise.}
    \label{fig:particle_ratio}
\end{figure}

\paragraph{\textbf{Noisy camera pose}} We implemented random variation in the camera pose estimates as the first noise addition to the simulations. The camera pose noise was included in all of the more complex simulations, as it was assumed that it appears the most consistently in any real-world application due to being the most dependent on real sensor noise. The noise only caused the models' estimates to converge slightly slower.

\paragraph{\textbf{False positive segments}} 
The incorrect segments were likely the largest cause of errors in the multi-view triangulation, as the model is reliant on estimating the centre point of each observation. With a distant target object, any false positive pixels greatly shifted the centre point of the segment. As seen in Table~\ref{tab:simple_results}, the false positives caused a significant jump in the simple multi-view triangulation errors. With a sufficiently small number of false positive frames as presented by our simulation, however, RANSAC can filter these erroneous frames and present a reliable target centre estimate through multi-view triangulation.

This type of noise was inherently easy to handle with the particle filter, as after convergence, any far false positives did not affect the filter updates. The biggest threat caused by false positives was during the initialisation step, where if a false positive appeared during one of the frames that were used for initialisation, the single error alone could cause the initial distribution to be off by kilometres. However, it could be accounted for by adjusting the size of the initial distribution. Another scenario where the false positives had a negative effect was when they appeared connected or only a few pixels away from a true positive target. In those situations, the filter would produce erroneous updates, but given that the amount of noise is sensible, these only cause the filter to produce momentary errors and converge slightly slower. 

\begin{figure*}
\centering
\newlength{\figheight}
\newlength{\figwidth}
\setlength{\figwidth}{0.2\textwidth}
\setlength{\figheight}{.5625\figwidth}
\begin{tikzpicture}
  \foreach \i in {1,2,3,4,5} {
    \node[minimum width=\figwidth,minimum height=\figheight] (t\i) at (\i*\figwidth,0) {\frame{\includegraphics[width=\figwidth]{figs/cropped_rw_og/rgb_\i.jpg}}};
    \node[minimum width=\figwidth,minimum height=\figheight] (b\i) at (\i*\figwidth,-\figheight) {\frame{\includegraphics[width=\figwidth]{figs/cropped_rw_og/seg_\i.jpg}}};
    \tikzmath{\k = (\i-1)*15;}
    }
    \node[anchor=south west,inner sep=4pt,font=\scriptsize\strut] at (b1.south west) {Masks for `mast'};
\end{tikzpicture}
    \centering
    \caption{The first empirical test sequence of the telecommunication mast target. On top are the drone-captured RGB images, and below are the used segments from edge detection. For visualisation, the segment has been dilated for an additional ten steps, and the images have been centre-cropped to half width and height.} 
    \label{fig:rw_rmse}
\end{figure*}

\begin{figure*}
\centering
\newlength{\figheightd}
\newlength{\figwidtd}
\setlength{\figwidtd}{0.2\textwidth}
\setlength{\figheightd}{.5625\figwidtd}
\begin{tikzpicture}
  \foreach \i in {1,2,3,4,5} {
    \node[minimum width=\figwidtd,minimum height=\figheightd] (t\i) at (\i*\figwidtd,0) {\frame{\includegraphics[width=\figwidtd]{figs/smoke_cropped/rgb_\i.jpg}}};
    \node[minimum width=\figwidtd,minimum height=\figheightd] (b\i) at (\i*\figwidtd,-\figheightd) {\frame{\includegraphics[width=\figwidtd]{figs/smoke_cropped/seg_\i.jpg}}};
    \tikzmath{\k = (\i-1)*15;}
    }
    \node[anchor=south west,inner sep=4pt,font=\scriptsize\strut] at (b1.south west) {Masks for `smoke'};

\end{tikzpicture}
    \centering
    \caption{The second empirical test sequence of the industrial smoke cloud target. On top, the drone-captured RGB images and below, the used segments from SAM 3. For visualisation, the images have been centre-cropped to half width and height.} 
    \label{fig:smoke_sequence}
\end{figure*}

\paragraph{\textbf{False negatives}} 
The incorrectly negative frames caused another significant jump in the simple multi-view triangulation performance. In situations with no false positives, they practically just reduce the number of usable frames, but when false positives are present, they cause the triangulation to be guided in an entirely wrong direction. RANSAC, however, is capable of filtering the errors caused by this type of noise as well. With the particle filter, the false positive frames simply slowed down the estimate convergence. 

\paragraph{\textbf{Partial false negatives}} 
The frames where only part of the true segment was removed did not present any major issues in the multi-view triangulation, even without RANSAC. However, they caused the most negative effects on the particle filter after the initialisation. In situations where part of the true positive pixels were covered for multiple consecutive frames, the filter started converging towards the wrong position, significantly reducing the target position estimation accuracy. This highlights that the model performs the best when used alongside a detection or segmentation model that can capture as many of the true positive pixels of the target object as possible. 

\subsection{Simulation Analysis and Model Optimisation}

Based on the simulation RMSEs, the multi-view triangulation and particle filtering provide similar estimates of the target centre position. 
The particle filter, however, has the advantage of modelling the target object structure, even in noisy scenarios, as highlighted by the consistent convergence of the particle target area ratio shown in the last column of Table~\ref{tab:simple_results} and Figure~\ref{fig:particle_ratio}.
Qualitative analysis of the particle filter convergence also showed that the distribution correctly models the direction of the greatest uncertainty. 
The distribution quickly dissipates from clear areas seen by the camera, but particles remain in plausible areas presented by the depth direction.

Through trial and error with various experiment configurations, we decided on a reliable set of default parameters. The chosen parameters were: step size $T_{s}=10~\text{m}$, number of particles $N_p=100\ 000$, number of no matches before filter dismissal $n_{\theta-dm}=5$, and single particle variance $\sigma_p=5~\text{m}$. These settings enabled the filters to locate the at different distances under realistic amounts of each type of noise. All the reported results were obtained using these parameters.

Each of the parameters is adjustable, and their effects are fairly intuitive. The dismissal parameter adjusts how the model behaves when targets appear or disappear, both intentionally and due to misdetections. With better-performing segmentation or detection models, the parameter can be set to lower values to obtain faster corrections, while in the presence of larger amounts of segmentation noise, the values should be higher to avoid generating false positive target locations or removing true target locations. With a smaller $T_{s}$, depending on the segmentation frame rate and camera translation speed, the thresholds can also be set higher, to correspond to a larger total translation for each behaviour. However, setting the $T_s$ value too small can negatively affect the convergence of the filter. 

\subsection{Empirical Data}

In the first communication mast experiment, the multi-view triangulation failed, mostly producing estimates with errors in the range of kilometres. The non-RANSAC version managed to get between 100 and 200 meters RMSE on the first few frames, but then shot up to thousands of meters due to the high number of false positives. The RANSAC version failed to find the correct position, even in those few frames, likely due to the lack of a sufficient number of inlier frames for a proper triangulation.
The particle filter was the only method that provided any sort of success, but with our proposed parameters, even the predictions of that model were around 300 metres RMSE after convergence. 
Seemingly, the model converged to a position that was significantly further than the actual target. 
Most likely, the problem was caused by particle sparsity and insufficient particle variance for the thin target object, but we did not further optimise the model for the specific sequence, and instead present results based on the model optimised on our simulation.

\begin{table}[!t]
    \centering
    \caption{Results from the second empirical experiment. The mean RMSE is computed from predictions where the camera translation from the first frame exceeds three-quarters of the sequence distance ($\sim$160~m).}
    \begin{tabular}{llrr}
    \toprule
       Experiment  & Method & RMSE min (m) & RMSE mean (m) \\
    \midrule
       Smoke & MVT & 179.09 & 210.20 \\
       Smoke & RMVT & 184.07 & 286.12 \\
       Smoke & PF & 226.84 & 358.17 \\ 
    \bottomrule
    \end{tabular}
    \label{tab:rw_results}
\end{table}

\begin{figure}[!t]
    \centering
    \includegraphics[width=0.98\linewidth]{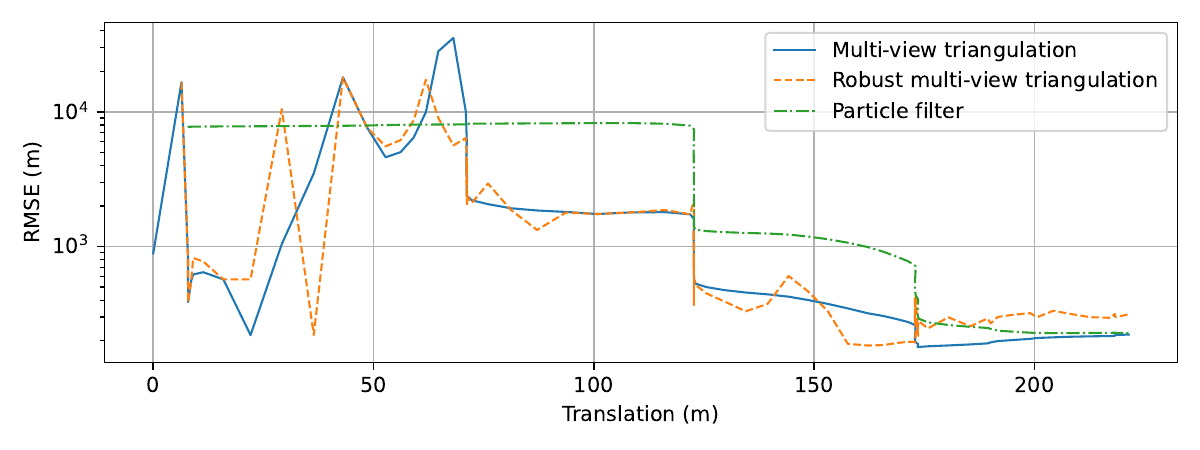}
    \caption{The evolution of localisation errors of different methods on the smoke sequence relative to the camera translation. The particle filter arrives at accurate estimates slightly later but presents smoother convergence.}
    \label{fig:smoke_error_plot}
\end{figure}

In the second test sequence, with the smoke cloud target, each model was able to localise the target object after approximately 150 to 180 metres of camera translation. 
Results after sufficient translation are collected in Table~\ref{tab:rw_results} and the evolution of the errors of different models' estimates over the camera translation are plotted in Figure~\ref{fig:smoke_error_plot}. 
Each model failed at very short translations ($<$50 metres), started to converge towards the right position after around 120 metres, and found a sufficient estimate near the end of the sequence ($\sim$180-230 metres). 
The sequence, albeit easier due to the larger target object and clearer segmentation, also represents a more realistic application scenario. 
We still considered the sequence challenging due to the noisy camera pose estimates and the small amount of camera translation relative to the target distance. 

In addition, we observed qualitatively that the particle distributions modelled the uncertainty correctly.
In the first experiment, the distribution always fell behind the mask, only failing to shift sufficiently in the depth direction. 
In the second experiment, the particles modelled the smoke cloud shape with, again, additional uncertainty in the direction from the camera to the target. 
Finally, the convergence of each model towards a similar position of 200 to 350 metre RMSE from the industrial chimney's position, used as the target location reference, shows that this region presents the most likely real centre point of the smoke cloud and that each tested model can be used to find the target centre location. 

\section{Conclusions}

Both the simulated and empirical tests showed that the task of localising distant objects in 3D using segments and known poses from a moving camera can be solved using either multi-view triangulation or a particle filter. 
The particle filter also models the target shape and the uncertainty of the resulting predictions, making it more reliable for practical applications. 

The results so far were limited to small offline datasets and simulations. Next steps for deploying the method on real applications require extending the simulation tests to an even larger variety of scenarios, including more real-world data testing, and implementing the algorithm on an embedded sensing system. In addition, the particle filter model could be extended to more complex multiple-target scenarios. The extension requires modelling situations such as scenarios with disappearing, fusing or separating targets.

Overall, the task presented in the study has very little representation in prior literature. 
This study shows that existing methods pose working solutions, and that the presented simulation method can be used to study the alternatives. Finally, the study implies that for the task of drone-based wildfire detection, the presented particle filter paired with a pre-existing segmentation model could solve the issue of finding wildfire geolocations at detection time.

\section{Acknowledgements}

We thank Teemu Hakala for setting up the instrumentation for the first empirical test data collection and Roope Näsi for conducting the first data-collection flights. We thank Arvi Solin for help with data collection for the second empirical test.
DALL-E 3 was used to sketch Figure~\ref{fig:teaser} in Section~\ref{sec:intro}.

\addtolength{\textheight}{-13.2cm} 

\bibliographystyle{ieeetr}
\bibliography{IEEEabrv,bib}

\end{document}